\documentclass[10pt,conference]{IEEEtran}
\IEEEoverridecommandlockouts

\usepackage{cite}
\usepackage{url}
\usepackage{amsmath,amssymb,amsfonts}
\usepackage{makecell}
\usepackage{algorithmic}
\usepackage{graphicx}
\usepackage{textcomp}
\usepackage{float}
\usepackage{xcolor}
\usepackage{booktabs}
\usepackage{multirow}
\usepackage{soul}
\def\BibTeX{{\rm B\kern-.05em{\sc i\kern-.025em b}\kern-.08em
    T\kern-.1667em\lower.7ex\hbox{E}\kern-.125emX}}
\begin{document}

\title{Improving Robustness of Vision-Language-Action Models by Restoring Corrupted Visual Inputs}


\author{
    \IEEEauthorblockN{
        Daniel Yezid Guarnizo Orjuela\IEEEauthorrefmark{2},
        Leonardo Scappatura\IEEEauthorrefmark{2},   
        Veronica Di Gennaro\IEEEauthorrefmark{2},
        \\ 
        Riccardo Andrea Izzo\IEEEauthorrefmark{3}, 
        Gianluca Bardaro\IEEEauthorrefmark{3}, and
            Matteo Matteucci\IEEEauthorrefmark{3}
    }
    \IEEEauthorblockA{
        Department of Electronics, Information, and Bioengineering\\
        Politecnico di Milano, Milano\\
        \IEEEauthorrefmark{2}(danielyezid.guarnizo, leonardo.scappatura, veronica.digennaro)@mail.polimi.it
        \\
        \IEEEauthorrefmark{3}(riccardo.izzo, gianluca.bardaro, matteo.matteucci)@polimi.it
    }
}

\maketitle

\begin{abstract}

Vision-Language-Action (VLA) models have emerged as a dominant paradigm for generalist robotic manipulation, unifying perception and control within a single end-to-end architecture. However, despite their success in controlled environments, reliable real-world deployment is severely hindered by their fragility to visual disturbances. While existing literature extensively addresses \textit{physical occlusions} caused by scene geometry, a critical mode remains largely unexplored: \textit{image corruptions}. These sensor-level artifacts, ranging from electronic noise and dead pixels to lens contaminants, directly compromise the integrity of the visual signal prior to interpretation. In this work, we quantify this vulnerability, demonstrating that state-of-the-art VLAs such as $\pi_{0.5}$ and SmolVLA, suffer catastrophic performance degradation, dropping from 90\% success rates to as low as 2\%, under common signal artifacts. To mitigate this, we introduce the \textbf{Corruption Restoration Transformer (CRT)}, a plug-and-play and model-agnostic vision transformer designed to immunize VLA models against sensor disturbances. Leveraging an adversarial training objective, CRT restores clean observations from corrupted inputs without requiring computationally expensive fine-tuning of the underlying model. Extensive experiments across the LIBERO and Meta-World benchmarks demonstrate that CRT effectively recovers lost performance, enabling VLAs to maintain near-baseline success rates, even under severe visual corruption.

\end{abstract}


\section{Introduction}
Recent advancements in robotic control tasks have been driven by Vision-Language-Action (VLA) models. By unifying perception, language understanding, and action prediction within a single architecture, these systems enable robots to execute complex manipulation tasks grounded in both visual observations and natural language instructions.

VLAs have the potential to bring significant advancements to numerous domains, ranging from industrial automation to the medical field. However, despite the impressive capabilities demonstrated in controlled research environments, widespread real-world deployment remains challenging due to significant reliability issues. These include satisfying real-time inference constraints while ensuring safety and robustness in dynamic and unstructured environments.

For a VLA to be deployable in the real world, it must exhibit \textit{Environmental Robustness} \cite{sapkota2025vision}, which refers to the capacity to maintain stable and accurate performance despite unpredictable variations in the input stream. Real-world environments are inherently complex, presenting challenges such as inconsistent lighting, adverse weather conditions, or instances where the target object is physically hidden by obstacles. 

Current literature actively addresses challenges arising from \textit{physical occlusions} in the scene, often resolving them through active perception or multi-view reasoning \cite{Wang2026PEAfowlPMA, liu2025trackvla++}. However, a distinct and equally critical source of visual degradation remains under-explored: \textit{image corruption}. Unlike physical obstacles, these are artifacts, ranging from water droplets and dead pixels to electronic interference, that directly compromise the integrity of the visual signal before it is interpreted by the VLA.
Existing studies indicate that standard VLAs, trained on clean, unobstructed visual data, exhibit significant performance degradation when key features are obscured, with a drop of 20–30\% under noisy conditions~\cite{sapkota2025vision}.

Nonetheless, a systematic quantification of how severely such image corruption artifacts impact the end-to-end success rates of modern VLAs is lacking in the current literature.

This paper addresses this gap through two contributions. First, we provide a rigorous quantitative analysis of performance degradation in state-of-the-art VLA models, specifically SmolVLA and $\pi_{0.5}$, under diverse image corruption scenarios. Our findings confirm that, without mitigation, these artifacts cause significant performance degeneration.

Second, we introduce the \textbf{Corruption Restoration Transformer (CRT)} to handle these vulnerabilities. We adopt the efficient Transformer-based architecture proposed by~\cite{verma2023image}, originally designed for image reconstruction, and repurpose it for the domain of robotics within high-resolution simulation environments. We designed it as a modular unit upstream of the VLA. This approach offers a strategic advantage: rather than retraining or fine-tuning from scratch a VLA, which would be computationally prohibitive, CRT acts as a specialized, lightweight module that restores observations before they reach the policy network.

The integration of CRT effectively minimizes divergence between actions derived from clean and corrupted inputs. effectively neutralizing the impact of image corruptions on the VLA. To facilitate reproducibility, we release our code, model and dataset\footnote{\url{https://huggingface.co/collections/AIRLab-POLIMI/corruption-restoration-transformer}}.
    
\section{Related Work}
\subsection{Vision-Language-Action Models (VLAs)}
Building upon the architectural foundations of Large Language Models (LLMs) and Vision-Language Models (VLMs), Vision-Language-Action (VLA) models extend these capabilities into the physical domain. This results in end-to-end systems, capable of reasoning and directly executing actions \cite{sapkota2025vision, cadene2024lerobot}.
VLAs are trained on internet-scale multimodal data paired with robot trajectories, such as the Open X-Embodiment dataset \cite{o2024open}.
Pioneering works, such as RT-2 \cite{zitkovich2023rt}, demonstrated that VLMs could be directly fine-tuned to output robotic actions by tokenizing continuous motions into discrete text tokens. This allows the model to leverage the intrinsic semantic reasoning of LLMs to enable zero-shot generalization to novel objects and instructions.
Building on this, OpenVLA \cite{kim2025openvla} introduced an open-source 7B-parameter architecture that integrates robust visual encoders (e.g., DINOv2 \cite{oquab2023dinov2} and SigLIP \cite{zhai2023sigmoid}) with a Llama-2 \cite{touvron2023llama} language backbone, establishing a robust baseline for generalist robot manipulation through discrete action token prediction.

Despite their success, treating continuous motor control as a discrete token prediction task often compromises the smoothness and precision required for manipulation. Consequently, recent architectures have transitioned towards diffusion-based or flow-matching heads to represent multimodal action distributions in continuous space. For instance, $\pi_0$ \cite{black2024pi0} employs flow-matching \cite{lipman2022flow} to predict continuous action chunks, while SmolVLA \cite{shukor2025smolvla} integrates cross-attention layers with a flow-matching action expert.
While these models excel in semantic generalization and task execution, their efficacy is highly dependent on the integrity of the visual encoder's feature space. Since they are primarily trained on curated datasets, their performance tends to degrade significantly when encountering visual corruptions, highlighting a critical lack of robustness.

\subsection{Visual Degradation in Robotic Perception}
Visual degradation poses a persistent challenge to reliable robotic control. We categorize these challenges into two classes based on their origin: \textit{Physical Occlusions} (e.g., scene-dependent) and \textit{Image Corruptions} (e.g., sensor-dependent).

\noindent\textbf{Physical Occlusion}: In unstructured real-world environments, target objects are frequently physically hidden behind other obstacles. The prevailing solution in the literature is active perception, where the robot autonomously adjusts its viewpoint to gather missing information. Recent frameworks such as VISO-Grasp \cite{shi2025viso} and Next-Best-View (NBV) planners \cite{jia2025pb} utilize VLMs to reason about spatial relationships and identify occluders, guiding the camera to a vantage point that reveals the target. Similarly, models like WoW \cite{chi2025wow} infer and remove obstructions through spatial reasoning. These methods rely on the premise that the obstructions are extrinsic; by changing the sensor's pose relative to the scene, the target can be revealed through multi-view fusion or scene repositioning.

\noindent\textbf{Image Corruption}: Unlike physical occlusions, sensor-based corruptions arise from artifacts inherent to the perception hardware or the signal transmission pipeline. These include water droplets, cracks, dead pixels, or electronic interference. While existing computer vision literature offers solutions for similar issues \cite{xu2026totnet, li2025seeing, mallick2025d}, these have rarely been adapted for end-to-end robotic systems. Crucially, standard techniques effective for physical occlusion are ineffective in this context. Active perception fails due to the corruption (e.g., a spot on the lens) being fixed in the camera frame and continuously obstructing a portion of the field of view regardless of the robot's position. Given that standard VLAs suffer performance drops of 20–30\% under these conditions \cite{sapkota2025vision}, there is a critical need for a modular restoration approach that cleanses the image observation before it reaches the policy network.
    
\section{Methodology}
\subsection{Overall Architecture}
\label{sec:overall_architecture}
To address the problem of image corruptions, we introduce a dedicated restoration module, the \textbf{Corruption Restoration Transformer (CRT)}, into the standard robotic control pipeline. Leveraging the LeRobot library~\cite{cadene2024lerobot}, we position the restoration module immediately upstream of the VLA model. In this configuration, every raw observation frame retrieved from the camera is processed to reduce corruptions before it is processed by the policy network

Let $x \in \mathbb{R}^{H \times W \times 3}$ be a clean observation frame from the robot camera. The spatial dimensions $H \times W$ are adapted to the simulator environment, with a resolution of $360 \times 360$ for the LIBERO~\cite{liu2023libero} benchmark and $480 \times 480$ for Meta-World~\cite{yu2020meta}.

\begin{figure}[t]
    \centering
    \includegraphics[width=1\linewidth]{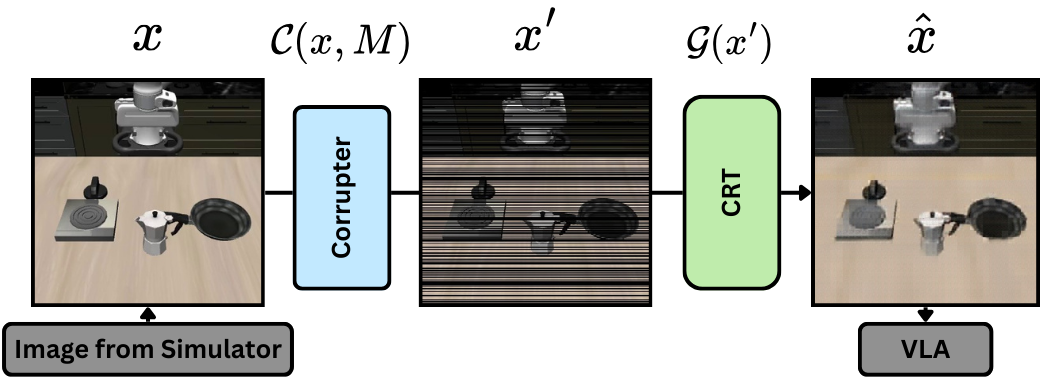}
    \caption{\textbf{System Pipeline.} The CRT module is inserted upstream of the VLA. It intercepts the corrupted simulator observation $x'$, restores it to $\hat{x}$, and passes the clean estimate to the policy network for action prediction.}
    \label{fig:pipeline} 
\end{figure}

We define a corruption process $\mathcal{C}$, such that the observed input is formulated as $x' = \mathcal{C}(x, M)$, where $M$ represents the specific artifact type applied to the clean frame. As illustrated in Figure~\ref{fig:pipeline}, the objective of CRT is to invert this process and recover the original signal $x$ from the corrupted input $x'$, ensuring that the VLA operates on a restored observation. 

\subsection{Corruption Restoration Transformer (CRT)}
We propose the Corruption Restoration Transformer (CRT), $\mathcal{G}$, a Transformer-based reconstruction model trained to map the corrupted image observation back to its original version. To achieve this, we draw inspiration from the image restoration framework introduced by~\cite{verma2023image}, and specialize the approach for the robotic perception domain. This architecture's ability to handle dense prediction tasks with high computational efficiency aligns with the needs of our target domain. We identify three critical architectural mechanisms to ensure that CRT provides the precision and efficiency required by the VLA:
\begin{itemize}
    \item \textbf{Shifted Patch Tokenization (SPT)}: Used to preserve spatial relationships between neighbouring pixels. SPT directly addresses the lack of local inductive bias in standard Vision Transformers (ViT) \cite{ferdous2024spt}.

    \item \textbf{Rotary Position Embeddings (RoPE)}: Adopted to encode both absolute and relative positions, allowing the model to generalize across longer action sequences used in action prediction \cite{su2024roformer}. 

    \item \textbf{Locality Self-Attention (LSA)}: Employed to sharpen attention scores and focus on local texture details.
\end{itemize}

The baseline architecture from~\cite{verma2023image} was originally designed for low-resolution ($64\times64$ pixels) tasks, whereas our application requires processing higher resolution visual inputs (up to $480\times480$ pixels) that contain dense semantic information. Therefore, it was necessary to significantly scale the model capacity to meet the requirements of the VLA. We increased the latent vector size to preserve fine-grained textures and object details critical for robotic manipulation with VLAs, but easily removed by corruption artifacts. Moreover, we expanded the number of transformer blocks, increasing the number of stacked encoder layers. In a Transformer, earlier layers typically process local patterns (edges, textures), while deeper layers aggregate this information into global semantic concepts. This increased depth provides the necessary capacity to model the complex statistics of the corruption artifacts over the whole image. Finally, to complement the increased depth, we also scaled the number of attention heads. This allows the model to reason about multiple relationships (e.g., distinguishing between a scene edge and a corruption line) in parallel.

\subsection{Adversarial Training}
\begin{figure}[t]
    \centering
    \includegraphics[width=1\linewidth]{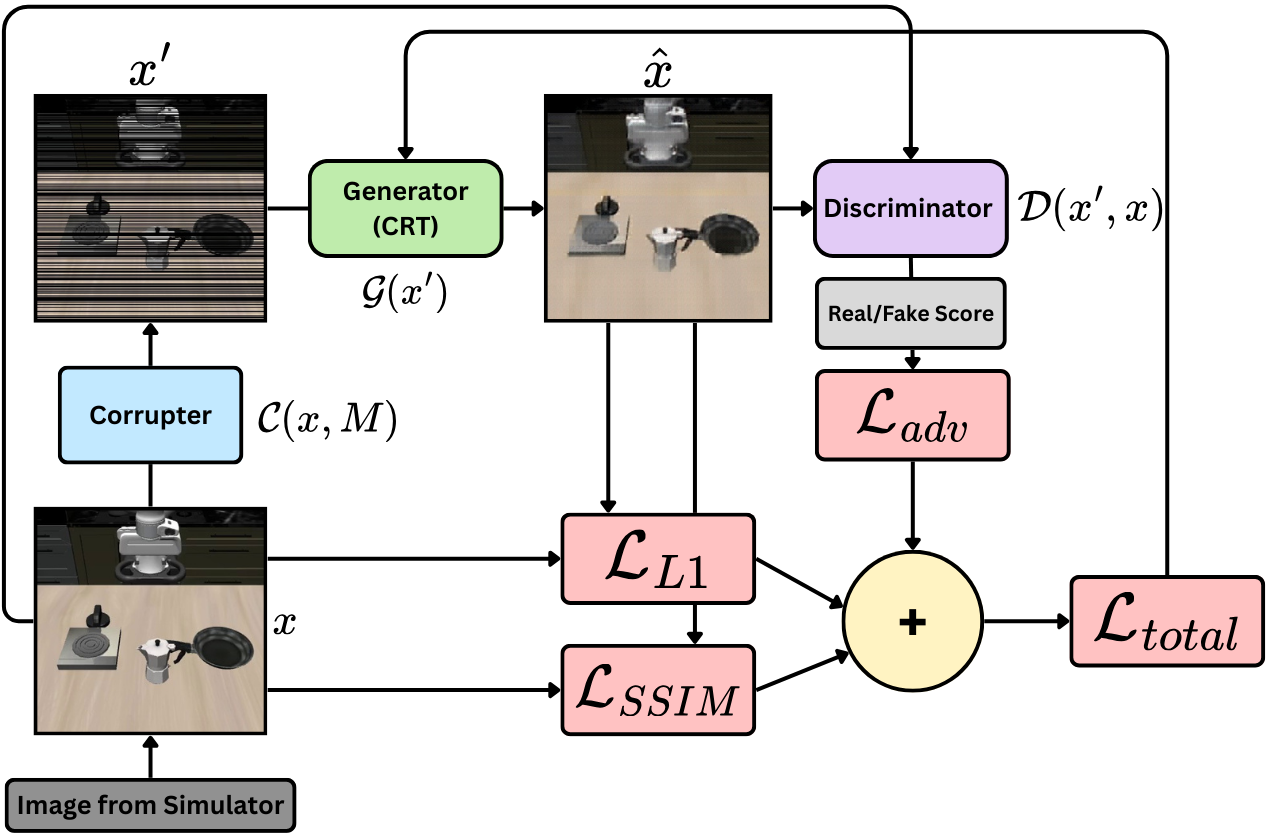} 
    \caption{\textbf{CRT Adversarial Training.} The CRT Generator ($\mathcal{G}$) receives a corrupted input $x'$ and produces a reconstructed observation $\hat{x}$. The network is optimized via a multi-objective loss function that combines pixel-wise fidelity $\mathcal{L}_{L1}$, structural similarity $\mathcal{L}_{SSIM}$, and adversarial feedback $\mathcal{L}_{adv}$ from the Discriminator $\mathcal{D}$ to ensure effective reconstructions.}
    \label{fig:adv_training}
\end{figure}

While standard pixel-wise reconstruction losses, such as $L_1$ or Mean Squared Error (MSE), are effective at recovering low-frequency global structures, they often result in blurry outputs when handling high-frequency details. In the context of robotic manipulation, preserving these details is critical for the downstream VLA to correctly interpret object affordances and boundaries. To mitigate this, we employ and extend the adversarial training framework suggested in~\cite{verma2023image}, treating the CRT module as a generator $\mathcal{G}$ within a GAN. Figure \ref{fig:adv_training} outlines our approach.

\begin{figure*}[t]
    \centering
    \includegraphics[width=1\textwidth]{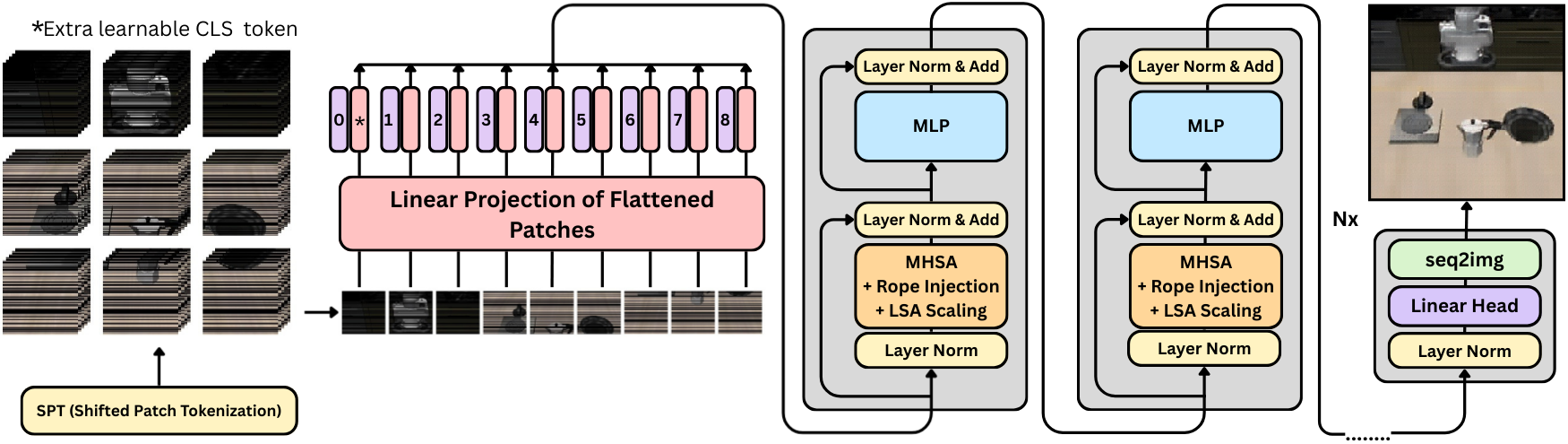}
    \caption{\textbf{Architecture of Corruption Restoration Transformer (CRT)} The model utilizes four specialized mechanisms: 1) Shifted Patch Tokenization (SPT) to recover local spatial dependencies; 2) Rotary Positional Embeddings (RoPE) for robust relative spatial encoding; 3) Locality Self-Attention (LSA) to sharpen textural details. The deep Transformer backbone enables the model to decouple complex corruption artifacts from the underlying semantic scene, finally reshaping the latent tokens into a restored RGB observation.}
    \label{fig:crt_architecture} 
\end{figure*}

\begin{figure}[t]
    \centering
    \includegraphics[width=1\linewidth]{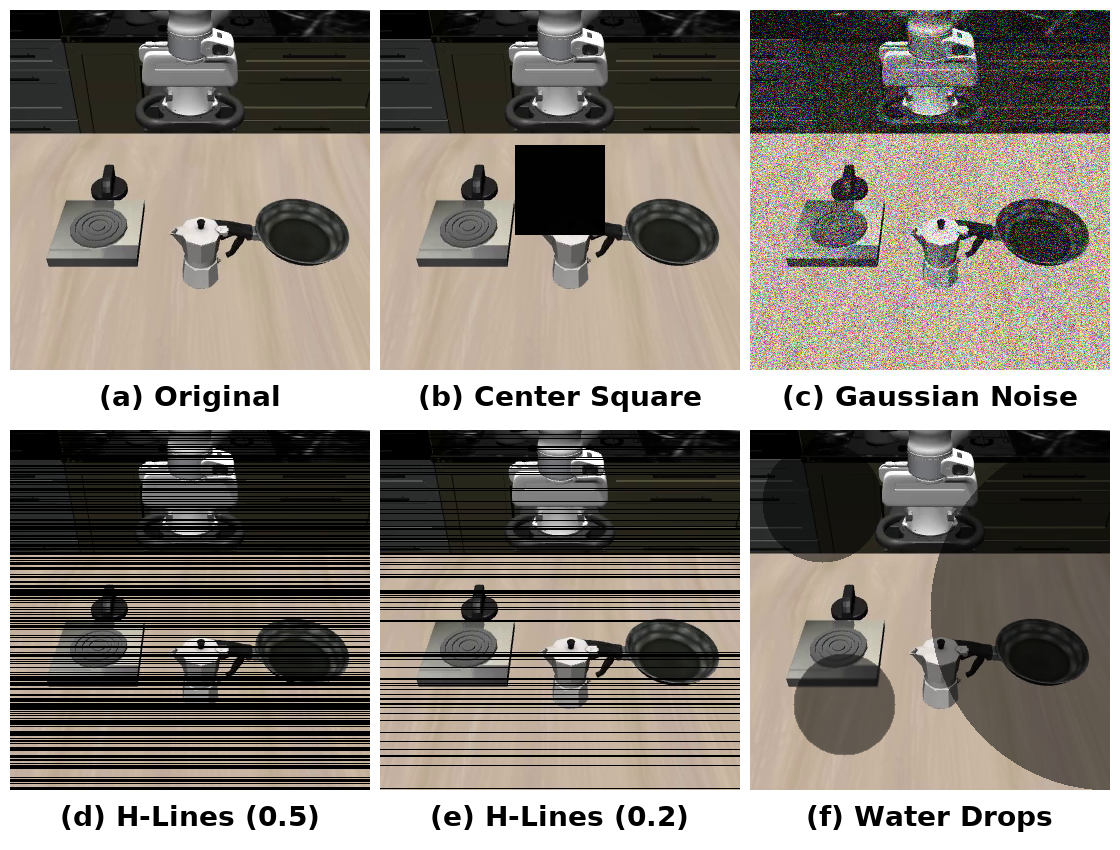}
    \caption{\textbf{Visual Corruption Types.} From top-left: (a) Clean Baseline, (b) Centered Square, (c) Gaussian Noise, (d) Horizontal Lines (0.5), (e) Horizontal Lines (0.2), (f) Water Drops.}
    \label{fig:corruptions}
\end{figure}

We introduce a discriminator network $\mathcal{D}$ trained to differentiate between ground-truth clean observations $x$ and the reconstructed observations $\hat{x} = \mathcal{G}(x')$. The discriminator architecture mirrors the transformer-based design of the generator but is optimized for classification. It processes inputs through a patch embedding layer and a sequence of Transformer blocks, with a final multi-layer perceptron (MLP) head to output a validity score. The training objective follows a multi-objective optimization strategy, utilizing a weighted combination of three distinct loss components:

\noindent\textbf{Adversarial Loss}: to encourage perceptual realism, we employ a standard binary cross-entropy objective. The discriminator aims to maximize the probability of correctly classifying real and reconstructed images, while the generator aims to minimize the probability that $\mathcal{D}$ recognizes $\hat{x}$ as fake:
\begin{equation}
    \mathcal{L}_{adv} = \mathbb{E}_{x}[\log \mathcal{D}(x)] + \mathbb{E}_{x'}[\log(1 - \mathcal{D}(\mathcal{G}(x')))]
\end{equation}

\noindent\textbf{Reconstruction Losses}: to ensure the reconstructed image remains faithful to the ground truth content, we utilize a pixel-wise $L_1$ loss, which promotes sparsity and reduces blurring compared to $L_2$ \cite{zhao2016loss}:
\begin{equation}
    \mathcal{L}_{L1} = \mathbb{E}_{x, x'} [\| x - \mathcal{G}(x') \|_1]
\end{equation}

\noindent\textbf{Structural Similarity Index Measure (SSIM) loss}\cite{brunet2011mathematical}: to preserve structural information such as brightness, contrast, and structure:
\begin{equation}
    \mathcal{L}_{SSIM} = 1 - \text{SSIM}(x, \mathcal{G}(x'))
\end{equation}

\noindent The final objective function for the generator $G$ is a weighted sum of these terms:
\begin{equation}
    \mathcal{L}_{total} = \lambda_{L1}\mathcal{L}_{L1} + \lambda_{SSIM}\mathcal{L}_{SSIM} + \lambda_{adv}\mathcal{L}_{adv}^{G}
\end{equation}
where $\mathcal{L}_{adv}^{G}$ represents the generator's component of the adversarial loss. Through empirical optimization, we set the hyperparameters to $\lambda_{L1}=10.0$, $\lambda_{SSIM}=1.0$, and $\lambda_{adv}=0.05$. The relatively low weight of the adversarial term ensures that the training remains grounded in content reconstruction while providing sufficient gradient signal to refine textural details and eliminate artifacts. 

\subsection{Dataset}
For the CRT training, we constructed two distinct paired datasets derived from successful execution trajectories within the LIBERO-10 and Meta-World MT50 benchmarks. We trained a dedicated reconstruction model for each benchmark, ensuring that CRT adapts to the distinct visual and semantic distributions of each simulation environment.  

To create the ground-truth reference dataset, we leveraged the pre-trained VLAs as expert demonstrators. For every task defined in the LIBERO-10 and Meta-World MT50 benchmarks, we executed the corresponding VLA policy in a noise-free environment. We strictly curated these rollouts, retaining only trajectories where the agent successfully achieved the goal state. To ensure a balanced distribution of visual semantics, we sampled exactly one successful trajectory for each distinct task (i.e., 10 for LIBERO, 50 for Meta-World). These trajectories were decomposed into individual frames to constitute the ground-truth observation set $x$, preserving the native simulation resolutions (i.e., $360 \times 360$ for LIBERO and $480 \times 480$ for Meta-World). Subsequently, for every extracted clean frame $x$, we applied the corruption generation process $\mathcal{C}(x, M)$ described in Section \ref{sec:image_corruptions}, to generate a corresponding corrupted $x'$.  This procedure resulted in perfectly aligned $(x', x)$ pairs suitable for supervised training.

The resulting datasets comprise 48,660 image pairs: 27,310 frames from LIBERO-10 and 21,350 from Meta-World MT50. Each dataset was partitioned into an 80/20 split for training and validation. We report training hyperparameters in Table \ref{tab:hyperparameters}.

\begin{table}[t]
\centering
\caption{\textbf{Training Hyperparameters.} We report the hyperparameters used to train CRT using a single NVIDIA RTX Quadro 6000.}
\label{tab:hyperparameters}
\begin{tabular}{@{}lcc@{}}
\toprule
\textbf{Parameter} & \textbf{LIBERO} & \textbf{Meta-World} \\ \midrule
Epochs & 30 & 37 \\
Learning Rate & 1e-4 & 7e-4 \\
Batch Size & 12 & 8 \\
Accumulation Steps & 12 & 32 \\
Training Time & 1d 8h 14m & 2d 22h 36m \\ \bottomrule
\end{tabular}
\end{table}

\section{Experiments and Results}
We evaluated the effectiveness of the CRT module using two widely recognized robotic manipulation benchmarks: LIBERO and Meta-World. These environments provide a diverse array of task complexities and semantic variations, while their integration with the LeRobot library ensures a standardized framework for reproducibility. We integrated our pipeline with two state-of-the-art VLAs: SmolVLA, evaluated on both LIBERO and Meta-World task suites, and $\pi_{0.5}$, evaluated exclusively on the LIBERO suite, as a fine-tuned checkpoint for Meta-World was unavailable at the time of the study.

\begin{table*}[t]
\caption{\textbf{Quantitative Analysis of CRT on LIBERO-10 and Meta-World MT50 Benchmarks.} Comparison of average success rates (SR) and relative performance degradation ($\Delta$) across clean and corrupted observations.}
\label{tab:main_results}
\centering
\resizebox{\textwidth}{!}{%
\begin{tabular}{ll|cc|cc|cc|cc|cc|cc}
\toprule
\multirow{2}{*}{\textbf{Benchmark}} & \multirow{2}{*}{\textbf{Model}} & \multicolumn{2}{c|}{\textbf{Baseline}} & \multicolumn{2}{c|}{\textbf{Centered Square}} & \multicolumn{2}{c|}{\textbf{Gaussian Noise}} & \multicolumn{2}{c|}{\textbf{Horiz. Lines (0.5)}} & \multicolumn{2}{c|}{\textbf{Horiz. Lines (0.2)}} & \multicolumn{2}{c}{\textbf{Water Drops}} \\
& & Avg. SR ($\uparrow$) & $\Delta$ & Avg. SR ($\uparrow$) & $\Delta$ & Avg. SR ($\uparrow$) & $\Delta$ & Avg. SR ($\uparrow$) & $\Delta$ & Avg. SR ($\uparrow$) & $\Delta$ & Avg. SR ($\uparrow$) & $\Delta$ \\
\midrule

\multirow{4}{*}{\textbf{LIBERO-10}} 
& $\pi_{0.5}$ & \textbf{90.0}\% & -- & 65.0\% & -27.78\% & \textbf{87.0}\% & -3.33\% & 2.0\% & -97.78\% & 62.0\% & -31.11\% & 68.0\% & -24.44\% \\
& $\pi_{0.5}$ + \textbf{CRT} & 89.0\% & -1.11\% & \textbf{89.0}\% & -1.11\% & 79.0\% & -12.22\% & \textbf{87.0}\% & -3.33\% & \textbf{89.0}\% & -1.11\% & \textbf{87.0}\% & -3.33\% \\
\cmidrule{2-14}
& SmolVLA & \textbf{43.0}\% & -- & \textbf{16.0}\% & -62.79\% & \textbf{13.0}\% & -69.77\% & 0.0\% & -100\% & 1.0\% & -97.67\% & 11.0\% & -74.42\% \\
& SmolVLA + \textbf{CRT} & 33.0\% & -23.26\% & 11.0\% & -74.42\% & 4.0\% & -90.70\% & \textbf{3.0}\% & -93.02\% & \textbf{15.0}\% & -65.12\% & \textbf{15.0}\% & -65.12\% \\
\midrule

\multirow{2}{*}{\textbf{Meta-World}} 
& SmolVLA & \textbf{58.0}\% & -- & 33.0\% & -43.10\% & 27.6\% & -52.41\% & 20.6\% & -64.48\% & 20.0\% & -65.52\% & 40.0\% & -31.03\% \\
& SmolVLA + \textbf{CRT} & 47.0\% & -18.96\% & \textbf{40.6}\% & -30.0\% & \textbf{34.6}\% & -40.34\% & \textbf{32.2}\% & -44.48\% & \textbf{41.4}\% & -28.62\% & \textbf{43.34}\% & -25.17\% \\

\bottomrule
\end{tabular}%
}
\end{table*}

\subsection{Image Corruptions}
\label{sec:image_corruptions}
To evaluate the robustness of VLA models against sensor-level disturbances, we selected a set of five distinct corruption types illustrated in Figure~\ref{fig:corruptions}. These artifacts are designed to model specific hardware failures, environmental conditions, and signal transmission errors commonly encountered in real-world robotic deployment.

\begin{itemize}
    \item \textbf{Centered Square}: In most manipulation tasks, the workspace and the main object of interest are centered in the frame. We simulate a catastrophic sensor failure that challenges the model by applying a black square to the image center, compromising the primary interaction zone.

    \item \textbf{Gaussian Noise}: This simulates electronic sensor noise typical of low-light conditions, high ISO settings, or thermal interference. We inject zero-mean Gaussian noise with a standard deviation of 0.20 into each pixel, degrading fine-grained visual features.

    \item \textbf{Horizontal Lines}: Represents scan-line artifacts from interlaced video transmission, rolling shutter effects, or electromagnetic interference. Black horizontal lines are randomly placed across the image, corrupting either 20\% (low intensity) or 50\% (high intensity) of the scene.

    \item \textbf{Water Drops}: Reproduces lens contamination from liquid splashes, condensation, or humidity. These conditions are frequently encountered in industrial or outdoor environments. Semi-transparent circular regions with Gaussian blur are applied at multiple random locations. 
\end{itemize}

\subsection{Results}
Our evaluation followed a three-stage protocol to validate the effectiveness of our CRT module across diverse manipulation tasks:
\begin{enumerate}
    \item We first evaluated SmolVLA and $\pi_{0.5}$ on clean, uncorrupted observations to establish a baseline and confirm the validity of the simulation environment.
    \item We then introduced the corruption process $\mathcal{C}(x, M)$, applying each of the five corruption types individually. We recorded the performance drop for each model without any mitigation strategies.
    \item Finally, we integrated the CRT module upstream of the VLA. We repeated the evaluation under the same corruption conditions to measure the system's ability to recover its performance.
\end{enumerate}

The outcomes of our evaluations are summarized in Table \ref{tab:main_results}. Our analysis focuses on three key dimensions: resilience to corruption of VLAs, CRT effectiveness, and the sensitivity of different VLAs on the restoration process.  

\noindent\textbf{Vulnerability of VLAs to Image Corruptions}: Consistent with previous literature, our results confirm that state-of-the-art VLA models are highly sensitive to sensor artifacts. As shown in Table \ref{tab:main_results}, performance collapses significantly under corruptions. For instance, $\pi_{0.5}$ on LIBERO-10 drops from a 90\% success rate to just 2\% under horizontal line corruption (0.5 intensity). Similarly, SmolVLA on Meta-World MT50 sees its performance halved from 58\% to 20.6\% under the identical conditions. These outcomes validate the premise that geometric generalization (e.g., active perception) is insufficient to mitigate cases when the underlying visual features are structurally compromised.

\noindent\textbf{Effectiveness of CRT}: The integration of CRT demonstrates a significant recovery, particularly for the $\pi_{0.5}$ architecture. As reported in Table~\ref{tab:main_results}, $\pi_{0.5}$ equipped with CRT achieves almost full recovery. Under the most severe ``Horizontal Lines (0.5)'' corruption, the success rate rebounds from 2\% to 87\%, effectively matching the clean baseline. Similar recovery is observed for ``Water Drops'' (68\% to 87\%) and ``Gaussian Noise'' (87\% to 79\%). Regarding SmolVLA on Meta-World, CRT consistently outperforms the unmitigated model across all corruption types. Notably, it improves success rates on ``Horizontal Lines'' by over 11\%, going from 20.6\% to 32.2\%, and shows strong resilience also to ``Centered Square'' occlusions, with an increment of 7.6\%.

\noindent\textbf{Baseline Preservation and Trade-offs}: A critical requirement for CRT is the preservation of performance on already clean observations.
For the $\pi_{0.5}$ model, CRT exhibits negligible overhead, maintaining an 89\% success rate on clean data compared to the 90\% baseline.
However, we observe a performance trade-off for the lighter SmolVLA architecture, where the baseline performance with CRT drops from 58\% to 47\% on Meta-World. This suggests that, while CRT successfully restores global semantic structure, smaller VLAs are less robust to feature perturbations and lack the capacity to generalize across subtle distributional shifts inherent to the reconstruction process. Nevertheless, this trade-off is significantly outweighed by the massive gains in reliability provided under adverse conditions. Furthermore, the CRT module is a lightweight neural network that introduces minimal overhead to the robotic control loop. Empirical tests indicated that CRT requires only 10-50 ms of inference time and approximately 1 GB of VRAM. In contrast, VLAs such as $\pi_{0}$ or SmolVLA require significantly higher computational resources with latencies in the order of hundreds of milliseconds for token generation. Consequently, CRT introduces a negligible overhead while increasing the robustness of the underlying model.

\section{Conclusion}

In this work, we identified a critical vulnerability in current Vision-Language-Action (VLA) models: their lack of robustness to image corruptions. While prior research has actively addressed physical occlusions through active perception, we demonstrated that sensor-level artifacts, such as lens contaminants, electronic noise, and signal loss, require a fundamentally different approach. Our systematic evaluation revealed that, even state-of-the-art models such as $\pi_{0.5}$ and SmolVLA, suffer catastrophic failure when subjected to these disturbances, with success rates dropping by nearly 90\% in severe cases. To bridge this gap, we introduced the Corruption Restoration Transformer (CRT), a specialized module designed to restore visual integrity upstream of the VLA. By leveraging advanced architectural mechanisms (e.g., SPT, RoPE, LSA) alongside an adversarial training objective, CRT effectively reconstructs clean observations from corrupted inputs, without requiring computationally prohibitive fine-tuning of the underlying VLA. Experimental results across LIBERO and Meta-World benchmarks confirm that CRT provides a robust solution against sensor artifacts. For large VLAs such as $\pi_{0.5}$, CRT acts as a near-perfect reconstructive layer, recovering performance to within 1-3\% of the clean baseline, even under heavy corruptions. We also observed that smaller architectures, such as SmolVLA, exhibit higher sensitivity to the subtle distributional shifts introduced by the reconstruction process, highlighting a trade-off between restoration effectiveness and input preservation. This limitation opens a clear avenue for future work. Given the modularity of CRT, the entire pipeline can be adapted to engage the restoration process only when a corruption is detected. Such an adjustment would maintain the original performance of the VLA, but also compensate for issues caused by image corruption. Ultimately, the CRT offers a scalable path towards more robust VLAs, deployable in the unpredictable and corrupted real-world environments.

\bibliographystyle{IEEEtran}
\bibliography{references}

\end{document}